\documentclass[10pt,twocolumn,letterpaper]{article}

\usepackage{wacv}
\usepackage{times}
\usepackage{epsfig}
\usepackage{graphicx}
\usepackage{amsmath}
\usepackage{amssymb}
\usepackage{bm}
\usepackage{dblfloatfix} 
\usepackage{enumitem}

\wacvfinalcopy 

\ifwacvfinal
\def\assignedStartPage{9876} 
\fi

\ifwacvfinal
\usepackage[breaklinks=true,bookmarks=false]{hyperref}
\else
\usepackage[pagebackref=true,breaklinks=true,colorlinks,bookmarks=false]{hyperref}
\fi

\ifwacvfinal
\else
\fi

\begin{document}

\title{Post-OCR Paragraph Recognition by Graph Convolutional Networks}

\author{Renshen Wang, Yasuhisa Fujii, Ashok Popat\\
Google Research\\
{\tt\small \{rewang, yasuhisaf, popat\}@google.com}
}

\maketitle

\begin{abstract}
   We propose a new approach for paragraph recognition in document images by spatial graph convolutional networks (GCN) applied on OCR text boxes. Two steps, namely line splitting and line clustering, are performed to extract paragraphs from the lines in OCR results. Each step uses a $\beta$-skeleton graph constructed from bounding boxes, where the graph edges provide efficient support for graph convolution operations. With pure layout input features, the GCN model size is 3$\sim$4 orders of magnitude smaller compared to R-CNN based models, while achieving comparable or better accuracies on PubLayNet and other datasets. Furthermore, the GCN models show good generalization from synthetic training data to real-world images, and good adaptivity for variable document styles.
\end{abstract}

\section{Introduction}

Document image understanding is a task to recognize, structure, and understand the contents of document images, and is a key technology to digitally process and consume such images, which are ubiquitous and can be found in numerous applications. Document image understanding enables the conversion of such documents into a digital format with rich structure and semantic information and makes them available for subsequent information tasks.

A document can be represented by its semantic structure and physical structure~\cite{Horak-Computer-1985}. The task to recover the semantic structure is called logical layout analysis~\cite{Cattoni-IRST-1998} or semantic structure extraction~\cite{Yang-CVPR-2017} while the task to recover the physical structure is called geometric (physical, or structural) layout analysis~\cite{Cattoni-IRST-1998}. These tasks are critical subproblems of document image understanding.

A paragraph is a semantic unit of writing consisting of one or more sentences that usually develops one main idea. Paragraphs are basic constituents of semantic structure and thus paragraph boundary estimation is an important building block of logical layout analysis. Moreover, paragraphs are often appropriate as processing units for various downstream tasks such as translation and information extraction because they are self-contained and have rich semantic information. Therefore, developing a generic paragraph estimation algorithm is of great interest by itself.

\begin{figure}[t]
\centering
\includegraphics[height=0.47\textwidth,angle=90]{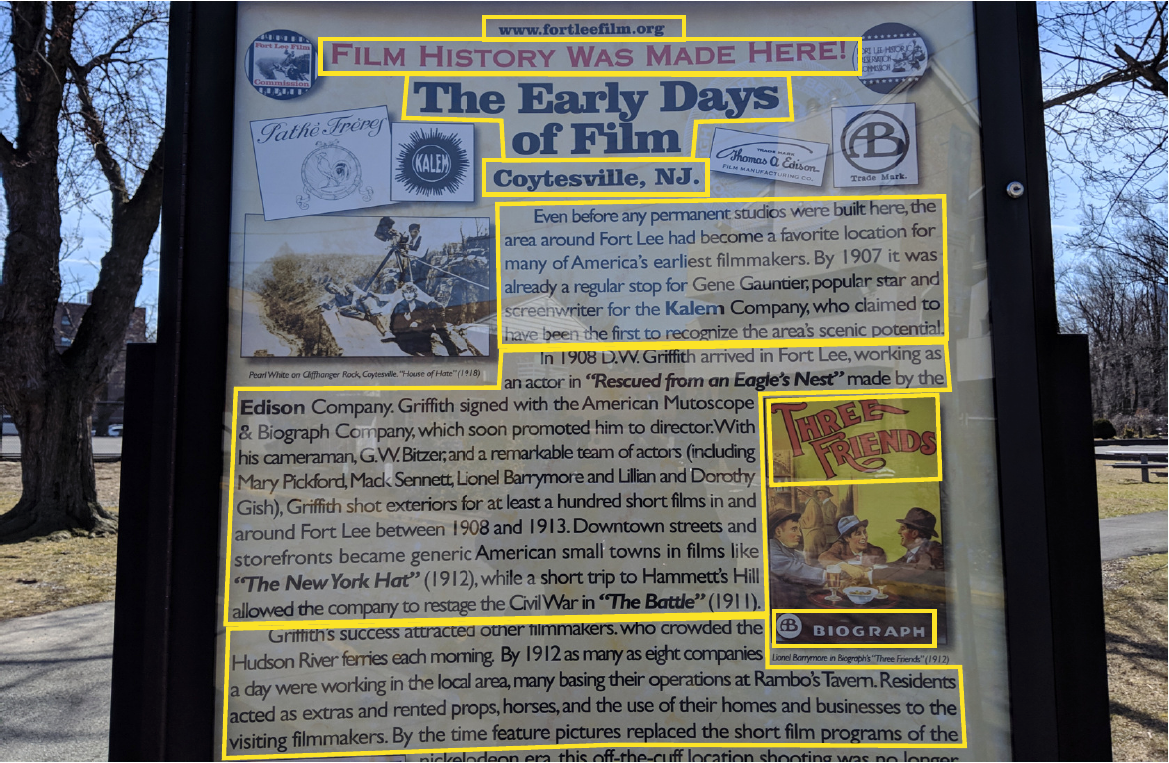}
\caption{Examples of paragraphs in printed text. Paragraphs may have complex shapes when wrapped around figures or other types of document entities.}
\label{fig:16}       
\end{figure}

Paragraphs are usually rendered in a geometric layout structure according to broadly accepted typographical rules. In this work, we exclude semantic paragraphs that can span over multiple text columns or pages, and only consider physical/geometrical paragraphs. There are usually clear visual cues to identify such paragraphs, but the task of estimating paragraphs is non-trivial as shown in Fig.~\ref{fig:16}.

Previous studies have attempted to develop a paragraph estimation method by defining handcrafted rules based on careful observations~\cite{10.5555/2887770.2887901,icpr1986,breuel-2003,DBLP:conf/icdar/Smith09} or by learning an object detection model to identify the regions of paragraphs from an image~\cite{Yang-CVPR-2017,DBLP:conf/icdar/ZhongTJ19}. For the former approaches, it is usually challenging to define a robust set of heuristics even for a limited domain, and hence machine-learning-based solutions are generally preferable. The latter approaches tend to have difficulty dealing with diverse aspect ratios and text shapes, and the wide range of degradations observed in real-world applications such as image skews and perspective distortions.

In this paper, we propose to apply graph convolutional networks (GCNs) in a post-processing step of an optical character recognition (OCR) system for paragraph recognition.
Recent advancements in graph neural (convolutional) networks~\cite{4700287,9046288} have enabled deep learning on non-Euclidian data. GCNs can learn spatial relationships among entities combining information from multiple sources and provide a natural way to learn the non-linear mapping from OCR results to paragraphs.

More specifically, we design two classifiers based on GCNs --- one for line splitting and one for line clustering. A word graph is constructed for the first stage and a line graph for the second stage. Both are constructed based on the $\beta$-skeleton algorithm \cite{BETA-skeleton} that produces a graph with good connectivity and sparsity.

To fully utilize the models' capability, it is desirable to have a diverse set of document styles in the training data. We create synthetic data sets from web pages where the page styles are randomly modified in the web scraping engine. By leveraging open web sites like Wikipedia~\cite{wiki} for source material to render in randomized styles, we have access to practically unlimited document data.

We evaluate the 2-step models on both the PubLayNet \cite{DBLP:conf/icdar/ZhongTJ19} and our own datasets. We show that GCN based models can be small and efficient by taking OCR produced bounding boxes as input, and are also capable of generating highly accurate results. Moreover, with synthesized training data from a browser-based rendering engine, these models can be a step towards a reverse rendering engine that recovers comprehensive layout structure from document images.

\section{Related Work}

\subsection{Page Segmentation}

A lot of previous work have studied the page segmentation task, including CRF based approaches \cite{icdar2007crf,das2014tao,icic2018li}, CNN based approaches \cite{Yang-CVPR-2017,DBLP:conf/icdar/LeeHOU19} and mixed algorithms \cite{DBLP:conf/sibgrapi/MaiaJH18}.

While the pixel masks from a segmentation can tell us where the paragraphs are, they do not produce individual paragraphs. For example, when text is dense and paragraphs are only hinted by subtle indentations, the adjacency graph in \cite{DBLP:conf/sibgrapi/MaiaJH18} produces many false positive edges that form multi-paragraph text components.

As a result, the problem we are trying to solve is different. Our work takes OCR result (text lines and words) rather than the image as input, and the goal is to recognize the paragraphs among the lines so as to improve the overall structure of the OCR engine output.

\subsection{Geometric and Rule-based Approaches}

Early studies have proposed geometric methods \cite{breuel-2003,1227629} and rule-based methods \cite{10.5555/2887770.2887901,icpr1986,DBLP:conf/icdar/Smith09}. Both categories have algorithms to find column gaps by searching for white space~\cite{1227629} or text alignment~\cite{DBLP:conf/icdar/Smith09}.

Limitations of these approaches include susceptibility to input noise and false positive column boundaries from monospace font families. Especially when handling scene text with perspective distortions from camera angles, rule based algorithms can be fragile and inconsistent.-eps-converted-to.pdf

\begin{figure}
\centering
\includegraphics[width=0.45\textwidth]{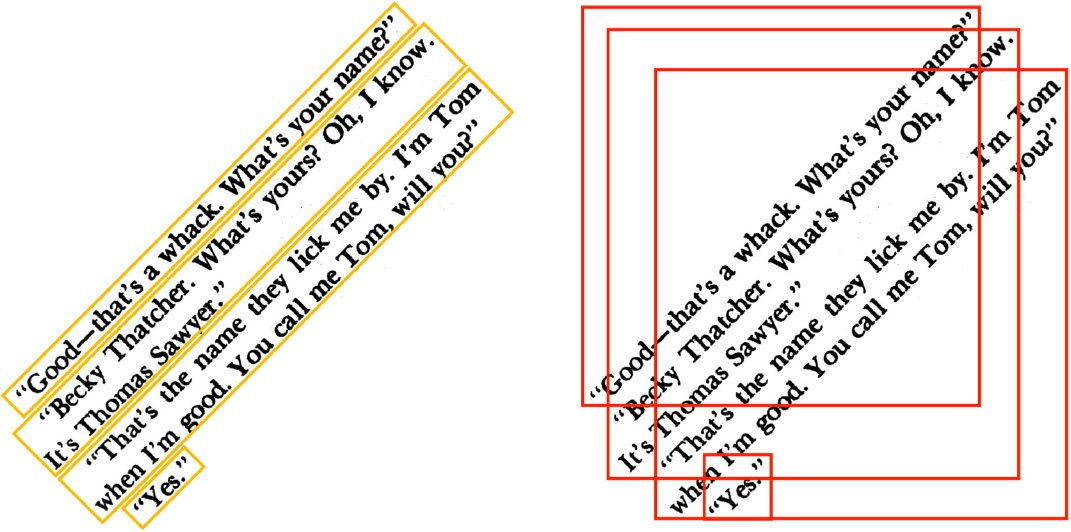}
\caption{Example of multiple short paragraphs densely packed and rotated into a non axis-aligned direction. The right side shows the region proposal boxes for object detection models.}
\label{fig:2}       
\end{figure}

\subsection{Image Based Detection}

The PubLayNet paper \cite{DBLP:conf/icdar/ZhongTJ19} provides a large dataset for multiple types of document entities, as well as two object detection models F-RCNN~\cite{DBLP:journals/pami/RenHG017} and M-RCNN~\cite{DBLP:journals/pami/HeGDG20} trained to detect these entities. Both show good metrics in evaluations, but with some inherent limitations.

\vspace{-2mm}
\begin{itemize}
  \itemsep0pt
  \item Cost: Object detection models are typically large in size and expensive in computation. When used together with an OCR engine to retrieve text paragraphs, it seems wasteful to bypass the OCR results and attempt to detect paragraphs independently.
  
  \item Quality: Paragraph bounding boxes may have high aspect ratios and are sometimes tightly packed. In Fig. \ref{fig:2}, several short paragraphs are printed with dense text and rotated by 45 degrees. The region proposals required to detect all the paragraphs are highly overlapped, so some detections will be dropped by non-maximum suppression (NMS). Rotational R-CNN models \cite{8545598} can mitigate this issue by inclined NMS, but further increase the computational cost while still facing a more difficult task with rotated or warped inputs.
\end{itemize}

\subsection{Graph Neural Networks}

Graph neural/convolutional networks have been used to extract document entities like tables~\cite{DBLP:conf/icdar/Riba0GFT019} and curved lines~\cite{cvpr2020zhang,pr2021ma}. 
These work show that graph neural networks are flexible for handling various types of entities with complex shapes. One possible limitation from these approaches is on graph construction -- the axis-aligned visibility graph in \cite{DBLP:conf/icdar/Riba0GFT019} can usually handle scanned documents but not scene text with image rotations and distortions, and the KNN graph in \cite{cvpr2020zhang,pr2021ma} can form isolated components that restrict graph operations. 

\section{Proposed Method}

\begin{figure}[t]
\includegraphics[height=0.48\textwidth,angle=90]{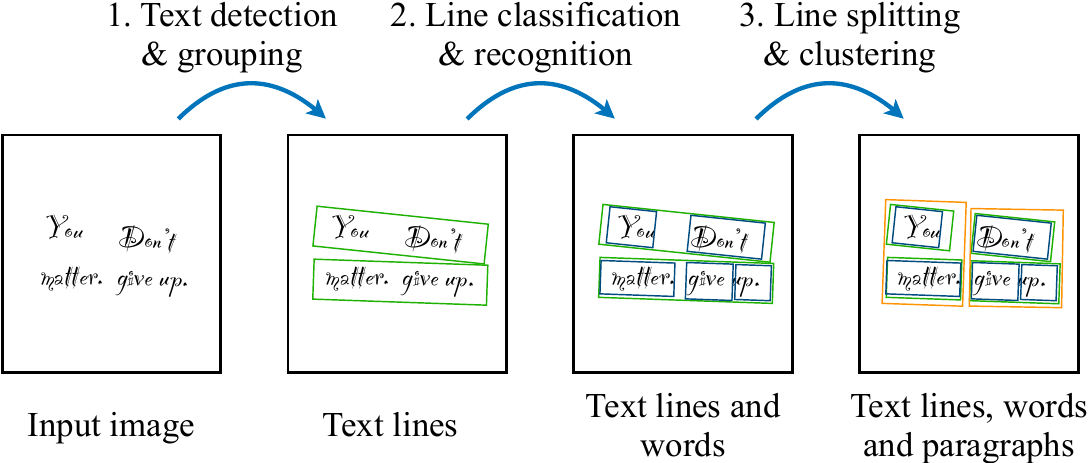}
\centering
\caption{Overview workflow of a typical OCR engine (stage 1 and 2) and our post-OCR paragraph recognition (stage 3). Lines are shown in green boxes, words in blue, and paragraphs in orange.}
\label{fig:17}       
\end{figure}

A typical general purpose OCR engine produces a set of text lines with recognized transcriptions \cite{das2018walker}. To find paragraphs, we can consider a bottom-up approach to cluster text lines into paragraphs.

As shown in Fig. \ref{fig:17}, the detected lines from stage 1 provide rudimentary layout information, but may not match the true text lines. The image in this example contains 2 text columns, each column containing a sentence which also forms a paragraph. The text line detector (stage 1) tries to find the longest curved fitted baselines, thus not able to split the lines by the 2-column layout. It is after stage 2 when the word boxes are available that we can perform a post-OCR layout analysis. We propose a 2-step process, namely line splitting and line clustering, to cluster the words and lines into paragraphs.

Both the line splitting and line clustering are non-trivial tasks for general-purpose paragraph estimation -- the input images can be skewed or warped, and the layout styles can vary among different types of documents, e.g. newspapers, books, signs, web pages, handwritten letters, etc. Even though the concept of paragraph is mostly consistent across all document categories, the appearance of a paragraph can differ by many factors such as word spacing, line spacing, indentation, text flowing around figures, etc. Such variations make it difficult, if not impossible, to have a straightforward algorithm that identifies all the paragraphs.

We design the two steps based on graph convolutional neural networks (GCN) \cite{9046288,NIPS2015_f9be311e} that takes input features from the coordinate values of OCR output boxes, together with a $\beta$-skeleton graph~\cite{BETA-skeleton} constructed from these boxes. Neither the original image nor text transcriptions are included in the input, so the models are small, fast, and entirely focused on the layout structure.

\vspace{-2mm}
\begin{itemize}
  \itemsep0pt
  \item Step 1: Line splitting. Raw text lines from OCR line detectors may cross multiple columns, and thus need to be split into shorter lines. A GCN node classifier takes word boxes to predict splitting points in lines.
  \item Step 2: Line clustering. The refined lines produced by step 1 are clustered into paragraphs. A GCN edge classifier takes line boxes to predict clustering operations on pairs of neighboring lines.
\end{itemize}
\vspace{-2mm}

Output of each model is applied to the OCR text lines, with some additional error-correction heuristics (e.g. lines too far apart should not be clustered).

\subsection{$\bm{\beta}$-skeleton on Boxes}
\label{beta_skeleton}

A graph is a key part of the GCN model input. We want a graph with high connectivity for effective message passing in graph convolutions, while also being sparse for computational efficiency.

\begin{figure}[b]
\includegraphics[height=0.48\textwidth,angle=90]{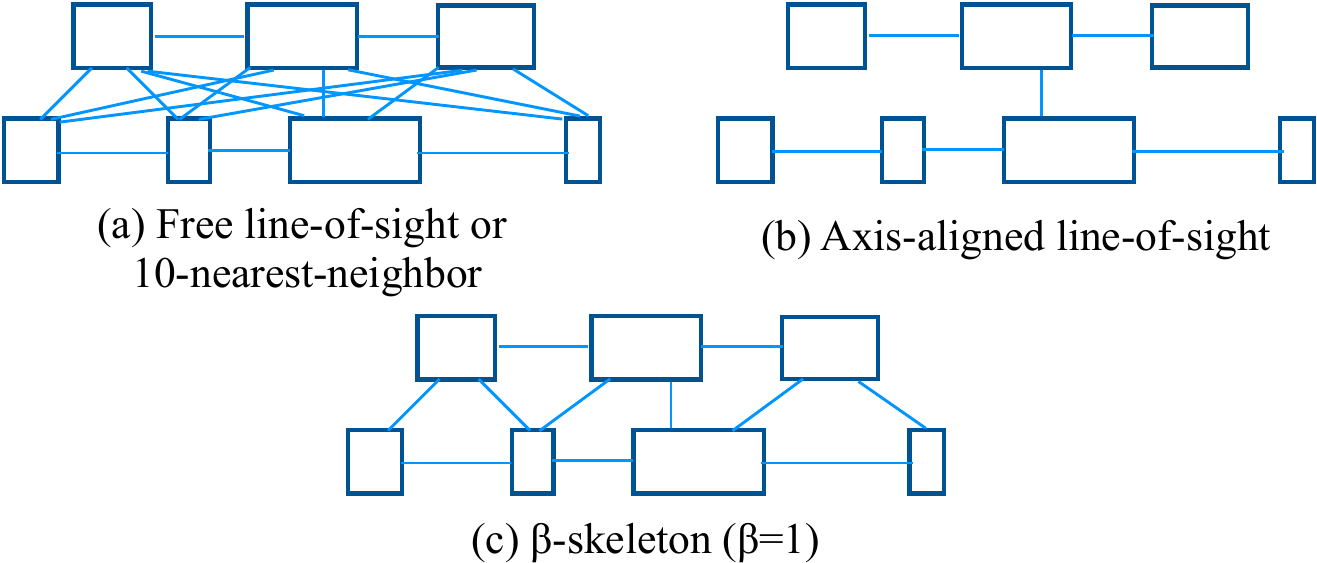}
\caption{Comparison among different types of graphs constructed on an example set of boxes.}
\label{fig:3}       
\end{figure}

\begin{figure}[b]
\centering
\includegraphics[height=0.45\textwidth,angle=90]{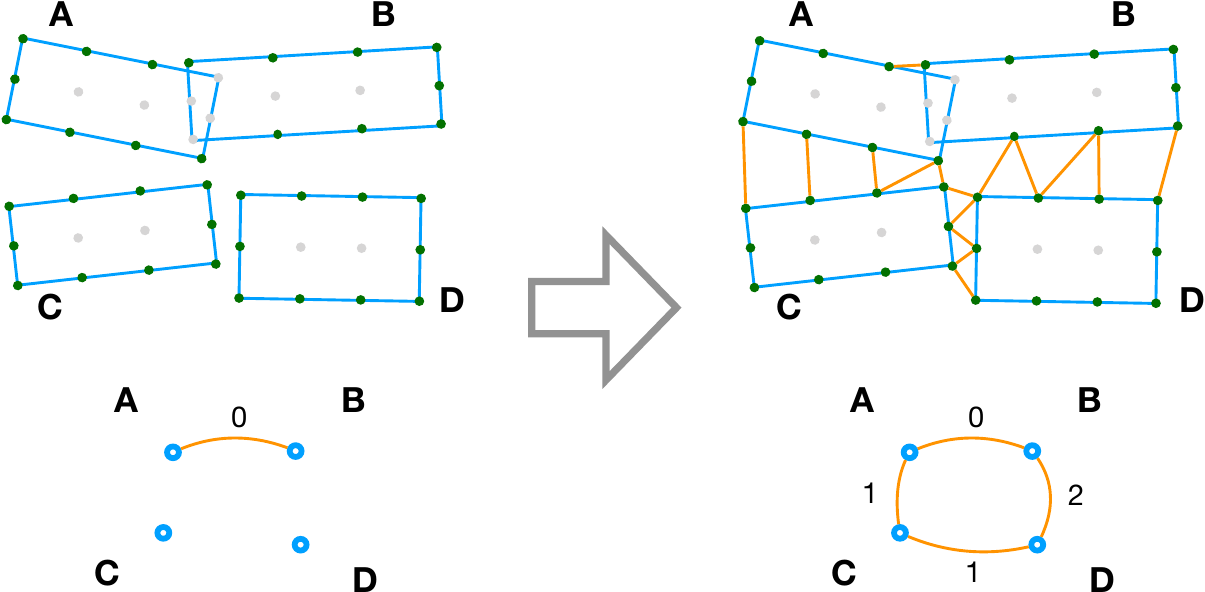}
\vspace{0.5mm}
\caption{Building a box $\beta$-skeleton from point based $\beta$-skeleton. Left side: intersecting boxes are first connected with edges of length 0. Right side: Non-internal peripheral points (in green) are connected with $\beta$-skeleton edges which are then collapsed into box edges. Edge lengths are approximate. The middle line points are added so that no edges can go through the boxes. }
\label{fig:4}       
\end{figure}

\begin{figure*}[b]
\centering
\includegraphics[height=0.85\textwidth,angle=90]{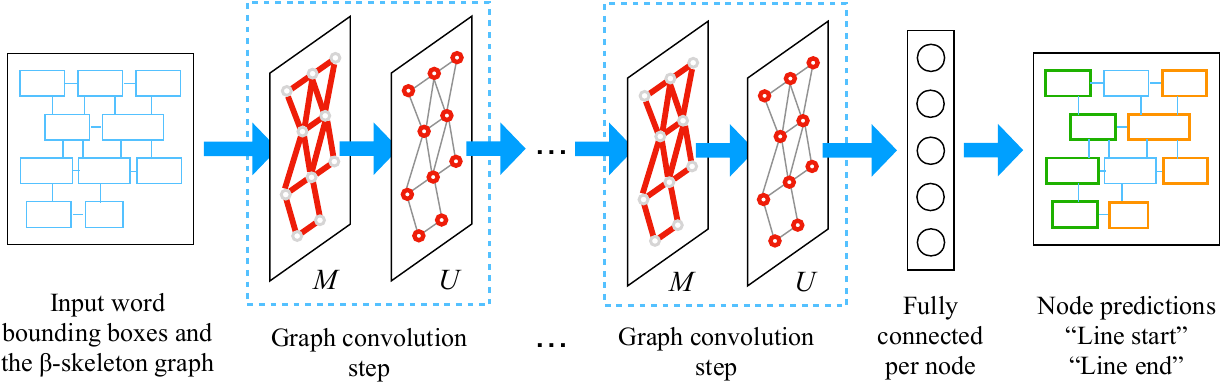}
\vspace{1mm}
\caption{Overview of the line splitting model. In the output, line start nodes are marked green and line end nodes are marked orange.}
\label{fig:5}       
\end{figure*}

Visibility graphs have been used in previous studies \cite{DBLP:conf/icdar/DavisMCPT19,DBLP:conf/icdar/Riba0GFT019}, where edges are made by ``lines-of-sight''. However, they are unsuitable for our models because of the edge density. Fig. \ref{fig:3}(a) shows  the visibility graph built on two rows of boxes, where any pairs of boxes on different rows are connected. This means word connections between text lines may get $O(n^2)$ number of edges. If we limit the lines-of-sight to be axis aligned like Fig. \ref{fig:3}(b), then the graph becomes too sparse, even producing disconnected components in some cases. In comparison, k-nearest-neighbor graphs used in \cite{cvpr2020zhang,pr2021ma} are more scalable, but can also produce dense and isolated components.

By changing ``lines-of-sight'' into ``balls-of-sight'' in visibility graphs, we get a $\beta$-skeleton graph \cite{BETA-skeleton} with $\beta = 1$. In such a graph, two boxes are connected if they can both touch a circle that does not intersect with any other boxes. It provides a good balance between connectivity and sparsity. As shown in Fig. \ref{fig:3}(c), a $\beta$-skeleton graph does not have excessive connections between rows of boxes. With $\beta = 1$, it is a subgraph of a Delaunay triangulation \cite{10.5555/1370949} with number of edges bounded by $O(n)$. Yet, it provides good connectivity within any local cluster of boxes, and the whole graph is guaranteed to be one connected component.

The original $\beta$-skeleton graph is defined on a point set. To apply it to rectangular boxes, we build a graph on peripheral points of all the box as in Fig. \ref{fig:4}, and keep at most one edge between each pair of boxes.

\subsection{Message Passing on Graphs}

\label{message_passing}

Our graph convolutional network is based on an early version of TF-GNN \cite{tfgnn} which works like MPNN \cite{10.5555/3305381.3305512} and GraphSage \cite{DBLP:conf/nips/HamiltonYL17}. We use the term ``message passing phase'' from \cite{10.5555/3305381.3305512} to describe the graph level operations in our models. In this phase, repeated steps of ``message passing'' are performed based on a message function $M$ and node update function $U$. At step $t$, a message $M(h_v^t, h_w^t)$ is passed along every edge $e_{vw}$ in the graph where $h_v^t$ and $h_w^t$ are the hidden states of node $v$ and $w$. Let $N(v)$ denote the neighbors of node $v$ in the graph, the aggregated message by average pooling received by $v$ is

\begin{equation}
m_v^{t+1} = \frac{\sum_{w \in N(v)} M(h_v^t, h_w^t)}{|N(v)|}
\end{equation}

\noindent and the updated hidden state is

\begin{equation}
h_v^{t+1} = U(h_v^t, m_v^{t+1})
\end{equation}

Alternatively, we can use attention weighted pooling \cite{DBLP:conf/iclr/VelickovicCCRLB18} to enhance message aggregation. Consequently, the model is also called a graph attention network (GAT), where calculation of $m_v^{t+1}$ is replaced by

\vspace{-1mm}
\begin{equation}
m_v^{t+1} = \frac{\sum_{w \in N(v)} \exp(e_{vw}^t) M(h_v^t, h_w^t)}{\sum_{w \in N(v)}\exp(e_{vw}^t)}
\end{equation}

\noindent and $e_{vw}^t$ is computed from a shared attention mechanism $a$, for which we use the dot product self-attention in \cite{DBLP:journals/corr/VaswaniSPUJGKP17}. So

\vspace{-1mm}
\begin{equation}
e_{vw}^t = a(h_v^t, h_w^t) = K(h_w^t) \cdot Q(h_v^t)
\end{equation}

\noindent where $K$ is a shared key function and $Q$ is a shared query function.

\begin{figure*}
\centering
\includegraphics[height=0.95\textwidth,angle=90]{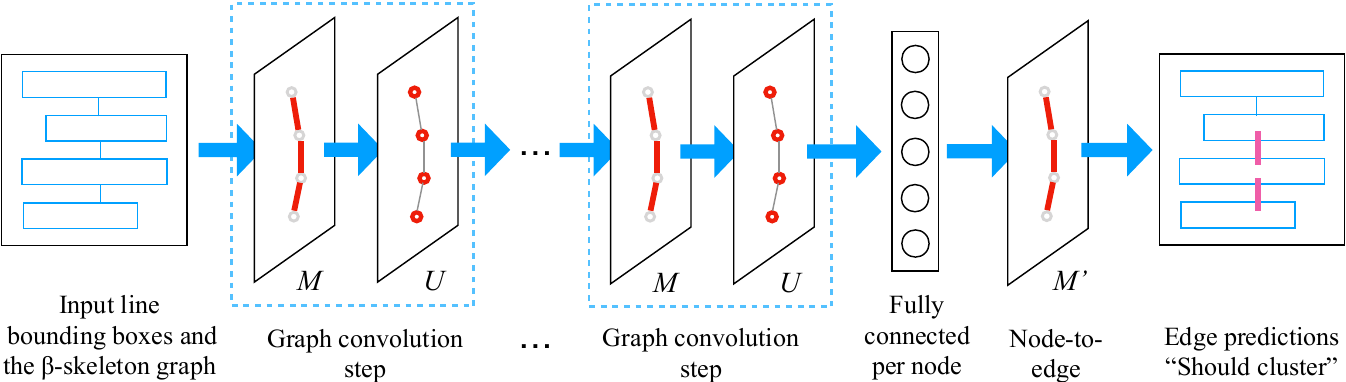}
\vspace{1mm}
\caption{Overview of the line clustering model. In the output, positive edges are marked pink.}
\label{fig:7}       
\end{figure*}

\subsection{Splitting Lines}

\label{splitting_lines}

\begin{figure}
\centering
\includegraphics[height=0.48\textwidth,angle=90]{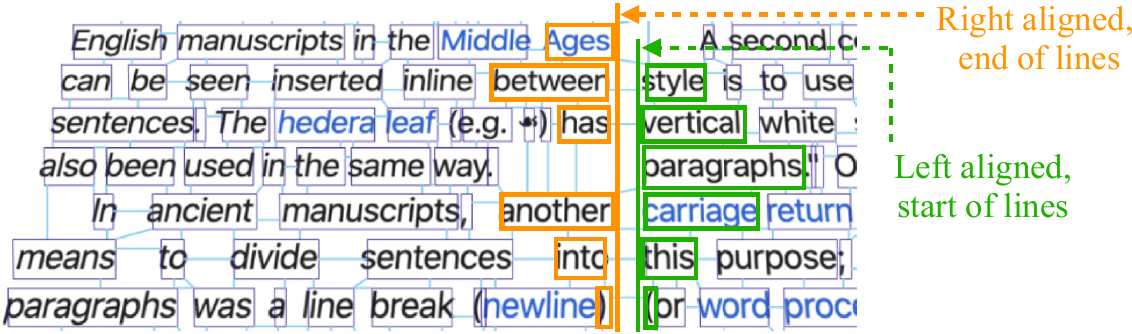}
\caption{Line splitting signal from word box alignment propagating through $\beta$-skeleton edges. The resulting predictions are equivalent to tab-stop detection.}
\label{fig:6}       
\end{figure}

When multi-column text blocks are present in a document page, splitting lines across columns is a necessary first step \cite{breuel-2003,DBLP:conf/icdar/Smith09}. Note that the horizontal spacings between words is not a reliable signal for this task, as when the typography alignment of the text is ``justified,'' i.e. the text falls flush with both sides, these word spacings may be stretched to fill the full column width. In Fig. \ref{fig:6}, the 2nd-to-last left line has word spacings larger than the column gap. This is common in documents with tightly packed text such as newspapers.

We use the GCN model shown in Fig. \ref{fig:5} to predict the splitting points, or tab-stops as in \cite{DBLP:conf/icdar/Smith09}. Each graph node is a word bounding box. Graph edges are the $\beta$-skeleton edges built as described in section \ref{beta_skeleton}. The model output contains two sets of node classification results -- whether each word is a ``line start'' and whether it is a ``line end''. 

Fig. \ref{fig:6} shows a $\beta$-skeleton graph constructed from the word bounding boxes. Since words are aligned on either side of the two text columns, a set of words with their left edges all aligned are likely on the left boundary of a column, i.e. these words are line starts. Similarly, a set of words with right edges aligned are likely on the right boundary, i.e. they are line ends. The $\beta$-skeleton edges can connect aligned words in neighboring lines, and pass box alignment signals for the effective learning of the GCN model. 

\subsection{Clustering Lines}

\label{clustering_lines}

After splitting all the lines into ``true'' lines, the remaining task is to cluster them into paragraphs. Again we use a GCN, but each graph node is a line bounding box, and the output is edge classification similar to link predictions in \cite{10.5555/1241540.1241551,DBLP:conf/nips/ZhangC18}. We define a positive edge to connect two consecutive lines in the same paragraph. Note that it is possible to have non-consecutive lines in the same paragraph being connected by a $\beta$-skeleton edge. Such edges are defined as negative to make the task easier to learn.

Fig. \ref{fig:7} is an overview of the line clustering model. The input consists of line bounding boxes, and an additional ``node-to-edge'' step is added for the final edge output:

\vspace{-1mm}
\begin{equation}
m'_{e=(v,w)} = \frac{M'(h_v, h_w) + M'(h_w, h_v)}{2}
\end{equation}

\begin{figure}
\centering
\includegraphics[height=0.48\textwidth,angle=90]{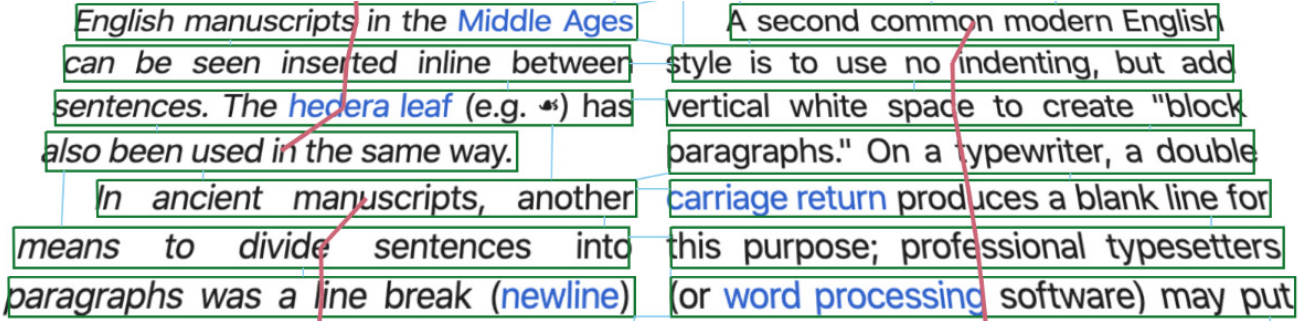}
\caption{Example of paragraph line clustering by indentations. Light blue edges indicate the $\beta$-skeleton constructed on line bounding boxes, and pink edges indicate that the connected lines are clustered into paragraphs.}
\label{fig:8}       
\end{figure}

The model predicts whether two lines belong to the same paragraph on each pair of lines connected with a $\beta$-skeleton edge. The predictions are made from multiple types of context like indentations (Fig. \ref{fig:8}) and line spacings. 

\section{Experiments}

We experiment with the 2-step GCN models and evaluate the end-to-end performance on both the open PubLayNet dataset, our synthetic web-scraped set, and a human annotated image set. The OCR engine is from Google Cloud Vision API DOCUMENT\_TEXT\_DETECTION v2021, and GCN setup details are in the appendix.

The GCN models are compared against other approaches. Besides the F-RCNN and M-RCNN from \cite{DBLP:conf/icdar/ZhongTJ19}, we train an F-RCNN model with additional quadrilateral outputs for rotated boxes, denoted by ``F-RCNN-Q'' in following subsections. It uses a ResNet-101~\cite{7780459} backbone at $\sim$200MB in size. In contrast, the GCN models are only $\sim$100KB each. A rule-based heuristic algorithm in our production system is also used as baseline.

\subsection{Datasets}

\subsubsection{PubLayNet}

PubLayNet~\cite{DBLP:conf/icdar/ZhongTJ19} contains a large amount of document images with ground truth annotations: 340K in the training set and 12K in the development/validation (dev) set. The testing set ground truth has not been released at the time of this writing, so we use the dev set for evaluation.

\subsubsection{Web Synthetic Page Layout}

Data diversity is a crucial necessity for handling all types of inputs. By taking advantage of high quality and publicly available web documents, as well as a powerful rendering engine used in modern browsers, we can generate synthetic training data with a web scraper.

We use a browser-based web scraper to retrieve a list of Wikipedia~\cite{wiki} pages, where each result includes the image rendered in the browser as well as the HTML DOM (document object model) tree. The DOM tree contains the complete document structure and detailed locations of all the rendered elements, from which we can reconstruct the ground truth line bounding boxes. Each line bounding box is an axis-aligned rectangle covering a line of text. For paragraph ground truth, the  HTML tag $<$p$>$ conveniently indicates a paragraph node, and all the text lines under this node belong to this paragraph.

\begin{figure}
\centering
\includegraphics[height=0.47\textwidth,angle=90]{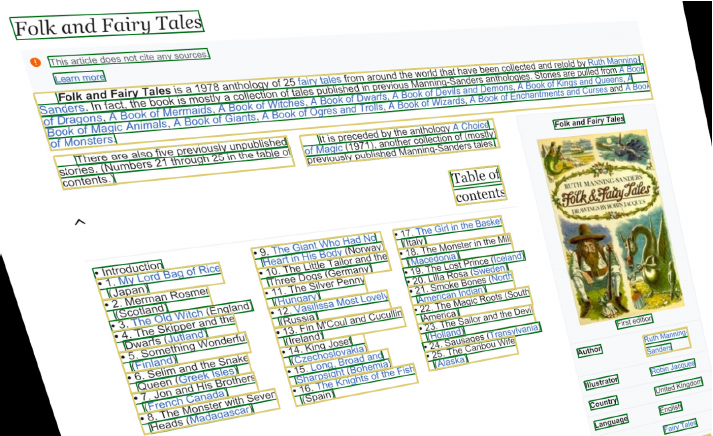}
\caption{Training data example from web scraping with randomized style changes and data augmentation. Green boxes indicate line ground truth labels and yellow boxes indicate multi-line paragraph ground truth labels.}
\label{fig:9}       
\end{figure}

By running extension scripts in the browser, we can randomly change layout styles of web pages and diversify our data. For example, to generate double-column text for a certain division of a page, we can use ``div.style.columnCount = 2.'' And to emulate the effect of camera angles on scene text, we further augment the data by perform randomized rotations and perspective projections on each scraped page. These methods produce a great variety of layout styles that can be encountered in the real word. Fig. \ref{fig:9} shows an example page from the augmented web synthetic data.

\subsubsection{Human Annotated Paragraph Dataset}

We have a human annotated set with real-world images -- 25K in English for training and a few hundred for testing in each available language. The images are collected from books, documents or objects with printed text, and sent to a team of raters who draw ground truth polygons for paragraphs. Example images are shown in Fig. \ref{fig:13}, \ref{fig:14} and \ref{fig:15}.

\subsection{Evaluation Metrics}

We measure the end-to-end performance of our OCR-GCN models by IoU based metrics such as the COCO mAP@IoU[.50:.95] used in \cite{DBLP:conf/icdar/ZhongTJ19} so the results are comparable. The average precision (AP) for mAP is usually calculated on a precision-recall curve. But since our models produce binary predictions, we have only one output set of paragraph bounding boxes, i.e. only one point on the precision-recall curve. So
$AP = precision \times recall$.

\begin{figure}
\centering
\includegraphics[height=0.45\textwidth,angle=90]{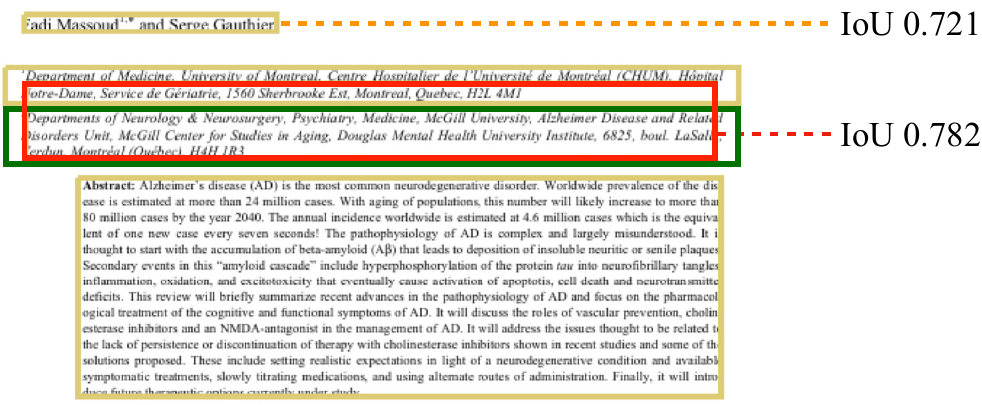}
\caption{A paragraph detection example on an image from PubLayNet~\cite{DBLP:conf/icdar/ZhongTJ19}. Yellow boxes are correct detections in terms of enclosed words, the red box is a wrong detection and the green box is ground truth. A single-line correct detection has lower IoU than a multi-line wrong one, which calls for F1$_{var}$ score with variable IoU thresholds. }
\label{fig:10}       
\end{figure}

For a better evaluation on paragraphs, we introduce an F1-score of variable IoU thresholds (F1$_{var}$ for short). As shown in Fig. \ref{fig:10}, a single-line paragraph has a lower IoU even though it is correctly detected, while a 4-line detection (in red) has a higher IoU with a missed line. This is caused by boundary errors at character scale rather than at paragraph scale. This error is larger for post-OCR methods since the OCR engine is not trained to fit the paragraph training data. If we have line-level ground truth in each paragraph, and adjust IoU thresholds $T_{iou}$ by
\begin{equation}
    T_{iou} = \min(1 - \frac{1}{1 + \#lines}, 0.95)
\end{equation}
the single-line paragraph will have IoU threshold 0.5, the 5-line one will have IoU threshold 0.833, and both cases in Fig. \ref{fig:10} can be more reasonably scored.

Both PubLayNet~\cite{DBLP:conf/icdar/ZhongTJ19} and our web synthetic dataset have line level ground truth to support this F1$_{var}$ metric. For the human annotated set without line annotations, we fall back to a fixed IoU threshold of 0.5.

\subsection{PubLayNet Evaluations}

The PubLayNet dataset has five types of layout elements: text, title, list, figure and table. For our task, we take text and title bounding boxes as paragraph ground truth, and set all other types as ``don't-care'' for both training and testing.

\begin{figure}
\centering
\includegraphics[height=0.47\textwidth,angle=90]{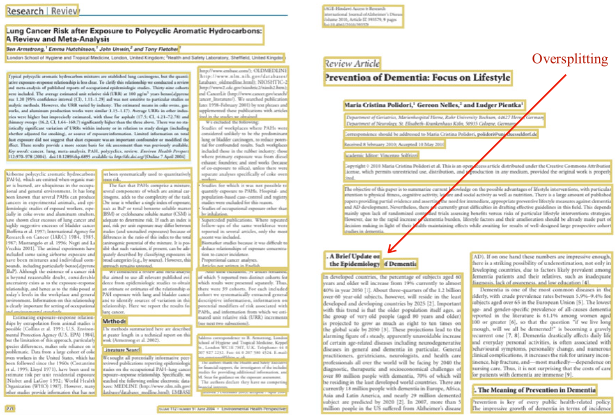}
\caption{Representative PubLayNet examples of paragraphs from OCR and GCN line splitting + clustering. The left one gets all the correct predictions, whereas the right one has a line splitting error at section 1's title.}
\label{fig:11}       
\end{figure}

Table \ref{tab:3} shows that F-RCNN-Q matches the mAP scores in \cite{DBLP:conf/icdar/ZhongTJ19}. The GCN models are worse in this metric because there is only one point in the precision-recall curve, and the OCR engine is not trained to produce bounding boxes that match the ground truth. In the bottom row of Table \ref{tab:3}, ``OCR + Ground Truth'' is computed by clustering OCR words into paragraphs based on ground truth boxes, which is the upper bound for all post-OCR methods. For mAP scores, even the upper bound is lower than the scores of image based models. However, if we measure by F1$_{var}$ scores defined above, OCR + GCNs can match image based models with a slight advantage. Fig. \ref{fig:11} shows some GCN produced examples. 

The high F1$_{var}$ score on ``OCR + Ground Truth'' also shows that the OCR engine we use has a very high recall on text detection. The only reason it is lower than one is from ground truth variations -- a small fraction of single-line paragraphs have IoU lower than 0.5.

\begin{table}
\footnotesize
\centering
\caption{Paragraph mAP@IoU[.50:.95] score and F1$_{var}$ score comparisons. All models are tested on the PubLayNet development set. Numbers for mAP in the first 2 rows are from~\cite{DBLP:conf/icdar/ZhongTJ19}.}
\label{tab:3}       
\begin{tabular}{lccc}
\noalign{\smallskip}\noalign{\smallskip}
\hline \hline\noalign{\smallskip}
Model & Training Set & mAP & F1$_{var}$ \\
\noalign{\smallskip}\hline\noalign{\smallskip}
F-RCNN~\cite{DBLP:conf/icdar/ZhongTJ19} & PubLayNet training & 0.910 & - \\
M-RCNN~\cite{DBLP:conf/icdar/ZhongTJ19} & PubLayNet training & 0.916 & - \\
F-RCNN-Q & PubLayNet training & 0.914 & 0.945 \\
Tesseract~\cite{Smith07anoverview} & - & 0.571 & 0.707 \\
OCR + Heuristic & - & 0.302 & 0.364 \\
OCR + GCNs & Augmented web synthetic & 0.748 & 0.867 \\
OCR + GCNs & PubLayNet training & 0.842 & 0.959 \\
OCR + Ground Truth \hspace{-2mm} & - & 0.892 & 0.997 \\
\noalign{\smallskip}\hline
\end{tabular}
\end{table}

\subsection{Web Synthetic Evaluations}

The synthetic dataset from web scraping gives a more difficult test for these models by its aggressive style variations. Data augmentation further increases the difficulty especially for image based detection models.

In Table \ref{tab:4}, we can see the F1$_{var}$ score of the image based F-RCNN-Q model decreases sharply as the task difficulty increases. At ``Augmented web synthetic'' with images like Fig. \ref{fig:9}, detection is essentially broken, not only from non-max suppression drops shown in Fig. \ref{fig:2}, but also from worse box predictions.

In contrast, the GCN models are much less affected by layout style variations and data augmentations. The F1$_{var}$ score change is minimal between augmented and non-augmented datasets. So GCN models will have a greater advantage for scene text when input images are rotated.

\begin{table}
\footnotesize
\centering
\caption{Paragraph F1$_{var}$ score comparison across different types of models and datasets. Data difficulty increases monotonically from PubLayNet to Augmented web synthetic. }
\label{tab:4}       
\begin{tabular}{lcc}
\noalign{\smallskip}\noalign{\smallskip}
\hline \hline\noalign{\smallskip}
Model & \ \ \ Data Source for Training \& Test \ \ \ \ \ & F1$_{var}$ \\
\noalign{\smallskip}\hline\noalign{\smallskip}
F-RCNN-Q & PubLayNet & 0.945 \\
& Web synthetic & 0.722 \\
& Augmented web synthetic \ & 0.547 \\
\noalign{\smallskip}\hline\noalign{\smallskip}
OCR + GCNs & PubLayNet & 0.959 \\
& Web synthetic & 0.830  \\
& Augmented web synthetic \ & 0.827 \\
\noalign{\smallskip}\hline
\end{tabular}
\end{table}

\begin{figure*}
\centering
\includegraphics[height=0.98\textwidth,angle=90]{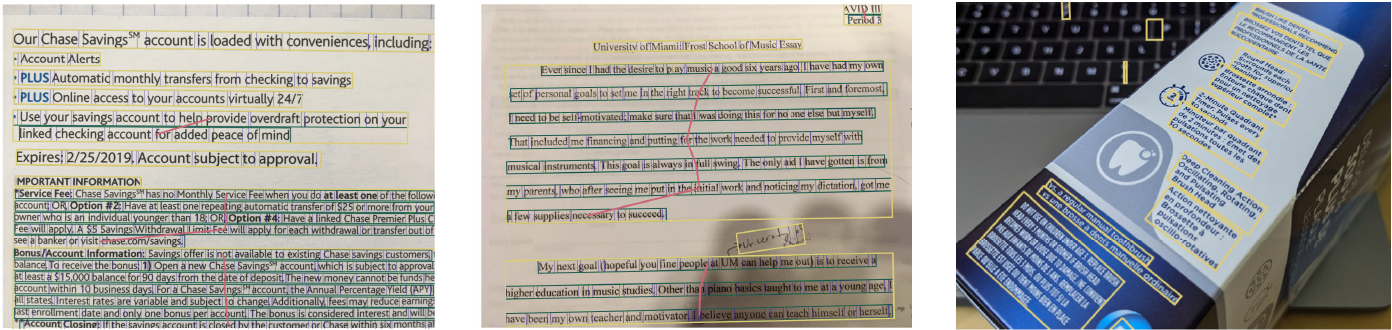}

\vspace{-1mm}
{\footnotesize (a) \hspace{52mm} (b) \hspace{52mm} (c) }
\caption{Representative examples of real-world images with OCR followed by GCN line splitting and line clustering. Paragraphs shown in yellow boxes, lines in green and words in blue. Pink line segments indicate positive line clustering predictions.}
\label{fig:13}       
\end{figure*}

\begin{figure*}
\centering
\includegraphics[height=0.98\textwidth,angle=90]{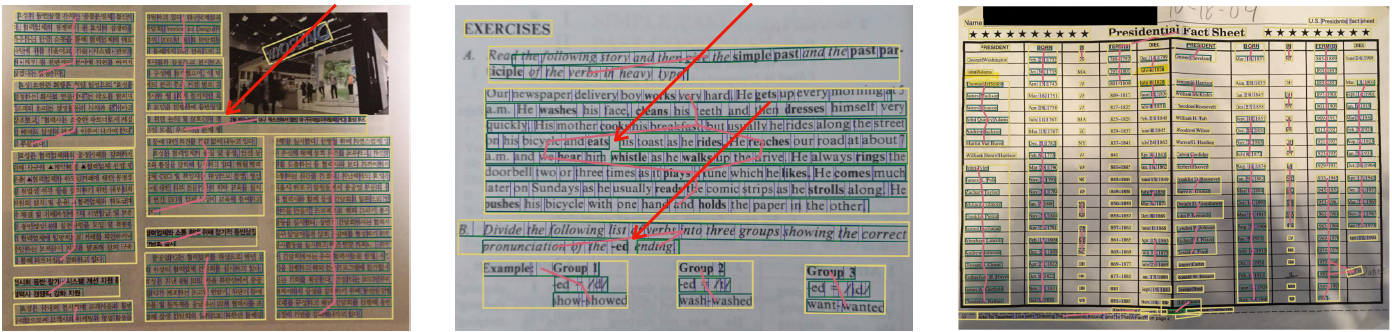}

\vspace{-1mm}
{\footnotesize (a) \hspace{52mm} (b) \hspace{52mm} (c) }
\caption{Paragraph errors in real-world images. (a) Line under splitting. (b) Line over splitting. (c) Over clustering table elements. }
\label{fig:14}       
\end{figure*}

\subsection{Real-world Dataset Evaluations}

The human annotated dataset can potentially show the models' performance in real-world applications. The annotated set is relatively small, so the F-RCNN-Q model needs to be pre-trained on PubLayNet, while the GCN models are small enough to be trained entirely on this set. Evaluation metric for this set is F1@IoU0.5.

Table \ref{tab:5} shows comparisons across different models and different training sets. Note that Faster R-CNN trained from synthetic web data does not work at all for real-world images, whereas the OCR+GCN models can generalize well. 

Fig. \ref{fig:13} and Fig. \ref{fig:14} show some examples of OCR + GCNs produced paragraphs. 
The right image in Fig. \ref{fig:13} shows the effectiveness of the augmented web synthetic data, as there are no similar images in the annotated set. On the other hand, the right table in Fig. \ref{fig:14} is not recognized since our models only takes bounding box coordinates as input. Using GCNs for table detection like \cite{DBLP:conf/icdar/Riba0GFT019} is another interesting topic but out of the scope of this paper.

\begin{table}
\footnotesize
\centering
\caption{Paragraph F1-scores tested on the real-world test set with paragraph annotations. Fixed IoU threshold 0.5 is used since there is no line-level ground truth to support variable thresholds.}
\label{tab:5}       
\begin{tabular}{llc}
\noalign{\smallskip}\noalign{\smallskip}
\hline \hline\noalign{\smallskip}
Model & \multicolumn{2}{r}{Training Data \hspace{20mm} F1@IoU0.5} \\
\noalign{\smallskip}\hline\noalign{\smallskip}
F-RCNN-Q & Augmented web synthetic & 0.030 \\
F-RCNN-Q & Annotated data (pre-trained & 0.607 \\
& on PubLayNet) \\
OCR + Heuristic & - &  0.602 \\
OCR + GCNs & Augmented web synthetic & 0.614 \\
OCR + GCNs & Annotated data & 0.671 \\
OCR + GCNs & Augmented synthetic + Annotated \ & 0.671 \\
OCR + Ground Truth & - & 0.960 \\
\noalign{\smallskip}\hline
\end{tabular}
\end{table}

\begin{table}[b]
\footnotesize
\centering
\caption{F1@IoU0.5 scores tested on the multi-language evaluation set.}
\label{tab:6}       
\begin{tabular}{lcccc}
\noalign{\smallskip}\noalign{\smallskip}
\hline \hline\noalign{\smallskip}
& OCR + & F-RCNN & OCR + & OCR + \\
Language & Heuristic & -Q & GCNs & Ground Truth \\
\noalign{\smallskip}\hline\noalign{\smallskip}
English & 0.429 & 0.513 & \textbf{0.544} & 0.890 \\
French & 0.438 & \textbf{0.557} & 0.553 & 0.885 \\
German & 0.427 & 0.538 & \textbf{0.566} & 0.873 \\
Italian & 0.455 & 0.545 & \textbf{0.556} & 0.862 \\
Spanish & 0.449 & 0.597 & \textbf{0.616} & 0.885 \\
\noalign{\smallskip}\hline\noalign{\smallskip}
Chinese & 0.370 & - & \textbf{0.485} & 0.790 \\
Japanese & 0.398 & - & \textbf{0.487} & 0.772 \\
Korean & 0.400 & - & \textbf{0.547} & 0.807 \\
\noalign{\smallskip}\hline
\end{tabular}
\end{table}

\begin{figure}
\includegraphics[height=0.475\textwidth,angle=90]{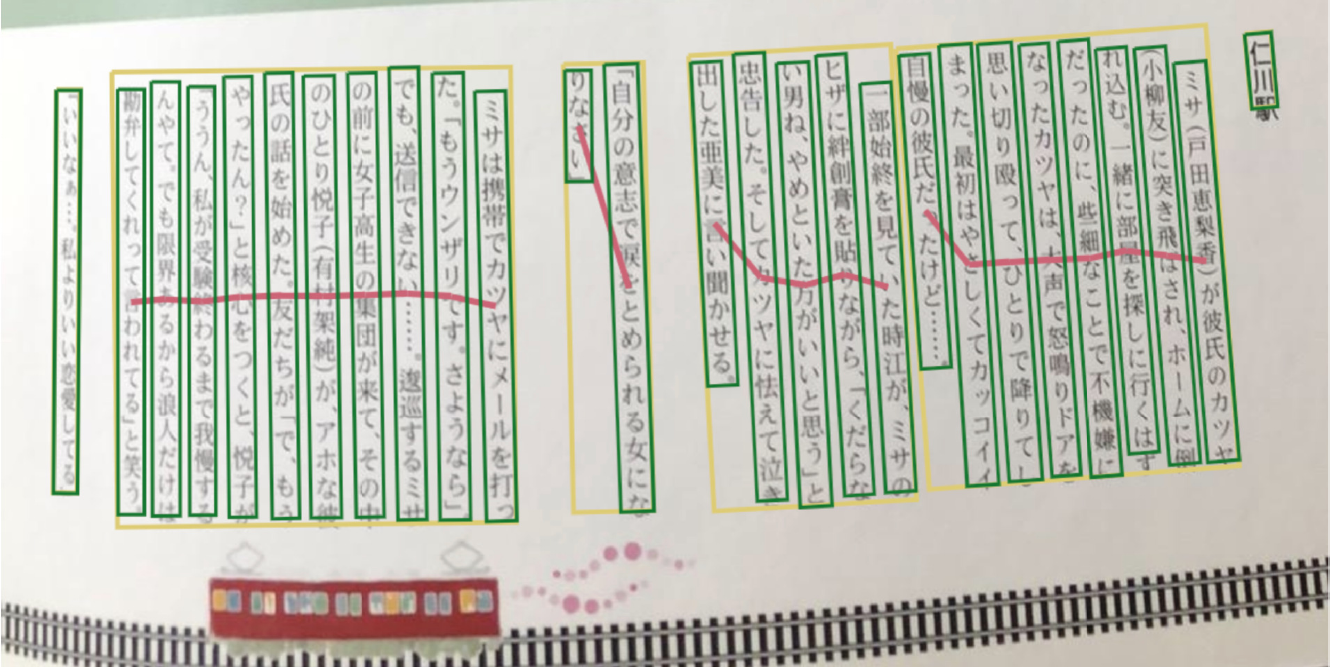}
\caption{Example of paragraphs from text lines with vertical writing direction.}
\label{fig:15}       
\end{figure}

To verify the robustness of the GCN models for language and script diversity, we test them on a multi-language evaluation set. The GCN models are trained with additional synthetic data from Wikipedia pages in Chinese, Japanese and Korean.
Table \ref{tab:6} once again shows the generalizability of GCN models. F-RCNN-Q is
not trained in the three Asian languages for the lack of training data.

The GCN models are also flexible in handling text lines written in vertical directions, which are common in Japanese and Chinese, and also appear in Korean. Although we don't have much training data with vertical lines, the bounding box structures of lines and symbols in these languages remain the same when the lines are written vertically, as if they were written horizontally while the image is rotated clockwise by 90 degrees. Fig. \ref{fig:15} shows such an example. Since our models are trained to handle all rotation angles, such paragraphs can be correctly recognized.

\section{Conclusions and Future Work}

We demonstrate that GCN models can be powerful and efficient for the task of paragraph recognition. Provided with a good OCR engine, they can match image based models with much lower requirement on training data and computation resources, and significantly beat them on non-axis-aligned inputs with complex layout styles. The graph convolutions in these models give them unique advantages in dealing with different levels of page elements and their structural relations.

Future work include extending the GCN models to find more types of entities and extract document structural information. Joining image based CNN backbones with GCN may work better for entities with non-text components like checkboxes and grid lines. In addition, reading order among entities will be helpful if we want to identify semantic paragraphs that span across multiple columns/pages. 
 
 \section*{Acknowledgment}

The authors would like to thank Chen-Yu Lee, Chun-Liang Li, Michalis Raptis, Sandeep Tata and Siyang Qin for their helpful reviews and feedback, and to thank Alessandro Bissacco, Hartwig Adam and Jake Walker for their general leadership support in the overall project effort.

{\small
\bibliographystyle{ieee_fullname}
\bibliography{egpaper}

\begin{thebibliography}{10}\itemsep=-1pt

\bibitem{wiki}
Wikipedia, the free encyclopedia.

\bibitem{10.5555/1370949}
Mark~de Berg, Otfried Cheong, Marc~van Kreveld, and Mark Overmars.
\newblock {\em Computational Geometry: Algorithms and Applications}.
\newblock Springer-Verlag TELOS, Santa Clara, CA, USA, 3rd ed. edition, 2008.

\bibitem{1227629}
T.~M. {Breuel}.
\newblock An algorithm for finding maximal whitespace rectangles at arbitrary
  orientations for document layout analysis.
\newblock In {\em Seventh International Conference on Document Analysis and
  Recognition, 2003. Proceedings.}, pages 66--70 vol.1, 2003.

\bibitem{breuel-2003}
Thomas~M. Breuel.
\newblock High performance document layout analysis.
\newblock In {\em Proceedings of the Symposium on Document Image Understanding
  Technology}, pages 209--218, Greenbelt, MD, 2003.

\bibitem{Cattoni-IRST-1998}
R. {Cattoni}, T. {Coianiz}, S. {Messelodi}, and C.~M. {Modena}.
\newblock Geometric layout analysis techniques for document image
  understanding: A review.
\newblock Technical Report 9703-09, IRST, Trento, Italy, 1998.

\bibitem{DBLP:conf/icdar/ClausnerAP19}
Christian Clausner, Apostolos Antonacopoulos, and Stefan Pletschacher.
\newblock {ICDAR2019} competition on recognition of documents with complex
  layouts - {RDCL2019}.
\newblock In {\em 2019 International Conference on Document Analysis and
  Recognition, {ICDAR} 2019, Sydney, Australia, September 20-25, 2019}, pages
  1521--1526. {IEEE}, 2019.

\bibitem{DBLP:conf/icdar/DavisMCPT19}
Brian~L. Davis, Bryan~S. Morse, Scott Cohen, Brian~L. Price, and Chris
  Tensmeyer.
\newblock Deep visual template-free form parsing.
\newblock In {\em 2019 International Conference on Document Analysis and
  Recognition, {ICDAR} 2019, Sydney, Australia, September 20-25, 2019}, pages
  134--141. {IEEE}, 2019.

\bibitem{NIPS2015_f9be311e}
David~K Duvenaud, Dougal Maclaurin, Jorge Iparraguirre, Rafael Bombarell,
  Timothy Hirzel, Alan Aspuru-Guzik, and Ryan~P Adams.
\newblock Convolutional networks on graphs for learning molecular fingerprints.
\newblock In C. Cortes, N. Lawrence, D. Lee, M. Sugiyama, and R. Garnett,
  editors, {\em Advances in Neural Information Processing Systems}, volume~28,
  pages 2224--2232. Curran Associates, Inc., 2015.

\bibitem{tfgnn}
Oleksandr Ferludin, Arno Eigenwillig, Martin Blais, Dustin Zelle, Jan Pfeifer,
  Alvaro Sanchez-Gonzalez, Sibon Li, Sami Abu-El-Haija, Peter Battaglia,
  Neslihan Bulut, Jonathan Halcrow, Filipe Miguel~Gonçalves de Almeida, Silvio
  Lattanzi, André Linhares, Brandon Mayer, Vahab Mirrokni, John Palowitch,
  Mihir Paradkar, Jennifer She, Anton Tsitsulin, Kevin Villela, Lisa Wang,
  David Wong, and Bryan Perozzi.
\newblock Tf-gnn: Graph neural networks in tensorflow, 2022.

\bibitem{10.5555/3305381.3305512}
Justin Gilmer, Samuel~S. Schoenholz, Patrick~F. Riley, Oriol Vinyals, and
  George~E. Dahl.
\newblock Neural message passing for quantum chemistry.
\newblock In {\em Proceedings of the 34th International Conference on Machine
  Learning - Volume 70}, ICML’17, page 1263–1272. JMLR.org, 2017.

\bibitem{DBLP:conf/nips/HamiltonYL17}
William~L. Hamilton, Zhitao Ying, and Jure Leskovec.
\newblock Inductive representation learning on large graphs.
\newblock In Isabelle Guyon, Ulrike von Luxburg, Samy Bengio, Hanna~M. Wallach,
  Rob Fergus, S.~V.~N. Vishwanathan, and Roman Garnett, editors, {\em Advances
  in Neural Information Processing Systems 30: Annual Conference on Neural
  Information Processing Systems 2017, 4-9 December 2017, Long Beach, CA,
  {USA}}, pages 1024--1034, 2017.

\bibitem{DBLP:journals/pami/HeGDG20}
Kaiming He, Georgia Gkioxari, Piotr Doll{\'{a}}r, and Ross~B. Girshick.
\newblock Mask {R-CNN}.
\newblock {\em {IEEE} Trans. Pattern Anal. Mach. Intell.}, 42(2):386--397,
  2020.

\bibitem{7780459}
K. {He}, X. {Zhang}, S. {Ren}, and J. {Sun}.
\newblock Deep residual learning for image recognition.
\newblock In {\em 2016 IEEE Conference on Computer Vision and Pattern
  Recognition (CVPR)}, pages 770--778, 2016.

\bibitem{Horak-Computer-1985}
W. {Horak}.
\newblock Office document architecture and office document interchange formats:
  Current status of international standardization.
\newblock {\em Computer}, 18:50--60, October 1985.

\bibitem{8545598}
Y. {Jiang}, X. {Zhu}, X. {Wang}, S. {Yang}, W. {Li}, H. {Wang}, P. {Fu}, and Z.
  {Luo}.
\newblock R2 cnn: Rotational region cnn for arbitrarily-oriented scene text
  detection.
\newblock In {\em 2018 24th International Conference on Pattern Recognition
  (ICPR)}, pages 3610--3615, 2018.

\bibitem{BETA-skeleton}
David~G. Kirkpatrick and John~D. Radke.
\newblock A framework for computational morphology.
\newblock {\em Machine Intelligence and Pattern Recognition}, 2:217--248, 1985.

\bibitem{726791}
Y. {Lecun}, L. {Bottou}, Y. {Bengio}, and P. {Haffner}.
\newblock Gradient-based learning applied to document recognition.
\newblock {\em Proceedings of the IEEE}, 86(11):2278--2324, 1998.

\bibitem{DBLP:conf/icdar/LeeHOU19}
Joonho Lee, Hideaki Hayashi, Wataru Ohyama, and Seiichi Uchida.
\newblock Page segmentation using a convolutional neural network with trainable
  co-occurrence features.
\newblock In {\em 2019 International Conference on Document Analysis and
  Recognition, {ICDAR} 2019, Sydney, Australia, September 20-25, 2019}, pages
  1023--1028. {IEEE}, 2019.

\bibitem{icic2018li}
Yixin Li, Yajun Zou, and Jinwen Ma.
\newblock Deeplayout: A semantic segmentation approach to page layout analysis.
\newblock In De-Shuang Huang, M.~Michael Gromiha, Kyungsook Han, and Abir
  Hussain, editors, {\em Intelligent Computing Methodologies}, pages 266--277,
  Cham, 2018. Springer International Publishing.

\bibitem{10.5555/1241540.1241551}
David Liben-Nowell and Jon Kleinberg.
\newblock The link-prediction problem for social networks.
\newblock {\em Journal of the American Society for Information Science and
  Technology}, 58(7):1019–1031, May 2007.

\bibitem{pr2021ma}
Chixiang Ma, Lei Sun, Zhuoyao Zhong, and Qiang Huo.
\newblock {ReLaText}: Exploiting visual relationships for arbitrary-shaped
  scene text detection with graph convolutional networks.
\newblock {\em Pattern Recognition}, 111:107684, mar 2021.

\bibitem{DBLP:conf/sibgrapi/MaiaJH18}
Ana Lucia Lima~Marreiros Maia, Frank~Dennis Julca{-}Aguilar, and Nina
  Sumiko~Tomita Hirata.
\newblock A machine learning approach for graph-based page segmentation.
\newblock In {\em 31st {SIBGRAPI} Conference on Graphics, Patterns and Images,
  {SIBGRAPI} 2018, Paran{\'{a}}, Brazil, October 29 - Nov. 1, 2018}, pages
  424--431. {IEEE} Computer Society, 2018.

\bibitem{icdar2007crf}
S. Nicolas, J. Dardenne, T. Paquet, and L. Heutte.
\newblock Document image segmentation using a 2d conditional random field
  model.
\newblock In {\em Ninth International Conference on Document Analysis and
  Recognition (ICDAR 2007)}, volume~1, pages 407--411, 2007.

\bibitem{10.5555/2887770.2887901}
Debashish Niyogi and Sargur~N. Srihari.
\newblock A rule-based system for document understanding.
\newblock In {\em Proceedings of the Fifth AAAI National Conference on
  Artificial Intelligence}, AAAI'86, page 789–793. AAAI Press, 1986.

\bibitem{DBLP:journals/pami/RenHG017}
Shaoqing Ren, Kaiming He, Ross~B. Girshick, and Jian Sun.
\newblock Faster {R-CNN:} towards real-time object detection with region
  proposal networks.
\newblock {\em {IEEE} Trans. Pattern Anal. Mach. Intell.}, 39(6):1137--1149,
  2017.

\bibitem{DBLP:conf/icdar/Riba0GFT019}
Pau Riba, Anjan Dutta, Lutz Goldmann, Alicia Forn{\'{e}}s, Oriol~Ramos
  Terrades, and Josep Llad{\'{o}}s.
\newblock Table detection in invoice documents by graph neural networks.
\newblock In {\em 2019 International Conference on Document Analysis and
  Recognition, {ICDAR} 2019, Sydney, Australia, September 20-25, 2019}, pages
  122--127. {IEEE}, 2019.

\bibitem{4700287}
F. {Scarselli}, M. {Gori}, A.~C. {Tsoi}, M. {Hagenbuchner}, and G.
  {Monfardini}.
\newblock The graph neural network model.
\newblock {\em IEEE Transactions on Neural Networks}, 20(1):61--80, 2009.

\bibitem{Smith07anoverview}
Ray Smith.
\newblock An overview of the {Tesseract OCR} engine.
\newblock In {\em Proc. 9th IEEE Intl. Conf. on Document Analysis and
  Recognition (ICDAR)}, pages 629--633, 2007.

\bibitem{DBLP:conf/icdar/Smith09}
Raymond~W. Smith.
\newblock Hybrid page layout analysis via tab-stop detection.
\newblock In {\em 10th International Conference on Document Analysis and
  Recognition, {ICDAR} 2009, Barcelona, Spain, 26-29 July 2009}, pages
  241--245. {IEEE} Computer Society, 2009.

\bibitem{icpr1986}
Sargur~N. Srihari and G.~W. Zack.
\newblock Document image analysis.
\newblock In {\em Proceedings of the 8th International Conference on Pattern
  Recognition}, page 434–436. AAAI Press, 1986.

\bibitem{das2014tao}
Xin Tao, Zhi Tang, Canhui Xu, and Yongtao Wang.
\newblock Logical labeling of fixed layout pdf documents using multiple
  contexts.
\newblock In {\em 2014 11th IAPR International Workshop on Document Analysis
  Systems}, pages 360--364, 2014.

\bibitem{DBLP:journals/corr/VaswaniSPUJGKP17}
Ashish Vaswani, Noam Shazeer, Niki Parmar, Jakob Uszkoreit, Llion Jones,
  Aidan~N. Gomez, Lukasz Kaiser, and Illia Polosukhin.
\newblock Attention is all you need.
\newblock {\em CoRR}, abs/1706.03762, 2017.

\bibitem{DBLP:conf/iclr/VelickovicCCRLB18}
Petar Velickovic, Guillem Cucurull, Arantxa Casanova, Adriana Romero, Pietro
  Li{\`{o}}, and Yoshua Bengio.
\newblock Graph attention networks.
\newblock In {\em 6th International Conference on Learning Representations,
  {ICLR} 2018, Vancouver, BC, Canada, April 30 - May 3, 2018}, 2018.

\bibitem{das2018walker}
Jake Walker, Yasuhisa Fujii, and Ashok Popat.
\newblock A web-based ocr service for documents.
\newblock In {\em 2018 13th IAPR International Workshop on Document Analysis
  Systems}, pages 21--22, 2018.

\bibitem{9046288}
Z. {Wu}, S. {Pan}, F. {Chen}, G. {Long}, C. {Zhang}, and P.~S. {Yu}.
\newblock A comprehensive survey on graph neural networks.
\newblock {\em IEEE Transactions on Neural Networks and Learning Systems},
  pages 1--21, 2020.

\bibitem{Yang-CVPR-2017}
Xiao {Yang}, Ersin {Yumer}, Paul {Asente}, Mike {Kraley}, Daniel {Kifer}, and
  C~Lee {Giles}.
\newblock Learning to extract semantic structure from documents using
  multimodal fully convolutional neural networks.
\newblock In {\em IEEE Conference on Computer Vision and Pattern Recognition
  (CVPR)}, 2017.

\bibitem{DBLP:conf/nips/ZhangC18}
Muhan Zhang and Yixin Chen.
\newblock Link prediction based on graph neural networks.
\newblock In {\em Advances in Neural Information Processing Systems 31: Annual
  Conference on Neural Information Processing Systems 2018, NeurIPS 2018, 3-8
  December 2018, Montr{\'{e}}al, Canada}, pages 5171--5181, 2018.

\bibitem{cvpr2020zhang}
Shi-Xue Zhang, Xiaobin Zhu, Jie-Bo Hou, Chang Liu, Chun Yang, Hongfa Wang, and
  Yin xu.
\newblock Deep relational reasoning graph network for arbitrary shape text
  detection.
\newblock pages 9696--9705, 06 2020.

\bibitem{DBLP:conf/icdar/ZhongTJ19}
Xu Zhong, Jianbin Tang, and Antonio Jimeno{-}Yepes.
\newblock Publaynet: Largest dataset ever for document layout analysis.
\newblock In {\em 2019 International Conference on Document Analysis and
  Recognition, {ICDAR} 2019, Sydney, Australia, September 20-25, 2019}, pages
  1015--1022. {IEEE}, 2019.

\end{thebibliography}
}

\nocite{726791}
\nocite{DBLP:conf/icdar/ClausnerAP19}

\clearpage

\appendix

\section{Algorithm to Construct the $\beta$-skeleton Graph on a Set of Boxes}

\renewcommand{\labelenumi}{\arabic{enumi}. }
\begin{enumerate}[leftmargin=0.22in]
  \itemsep0pt
  \item  For each box, pick a set of peripheral points at a pre-set density, and pick a set of internal points along the longitudinal middle line. 

  \item  Build a Delaunay triangulation graph $G_D$ on all the points. (Time complexity $O(n \log n)$ \cite{10.5555/1370949}.)

  \item  Find all the ``internal'' points that are inside at least one of the boxes. (Time complexity $O(n)$ by traversing along $G_D$'s edges inside each box starting from any peripheral point. Internal points marked grey in Fig. \ref{fig:4}.)

  \item Add an edge of length 0 for each pair of intersecting boxes (containing each other's peripheral points).
  
  \item Pick $\beta$-skeleton edges from $G_D$ where for each edge $e=(v_1, v_2)$, both its vertices $v_1$, $v_2$ are non-internal points and the circle with $v_1 v_2$ as diameter does not cover any other point.
  
  If there is such a point set $V_c$ covered by the circle, then the point $v_3 \in V_c$ closest to $\overline{v_1v_2}$ must be the neighbor of either $v_1$ or $v_2$ (in Delaunay triangulation graphs). Finding such $v_3$ takes $O(\log n)$ time for each edge, since $G_D$ produced in step 2 have edges sorted at each point.
    
  \item  Keep only the shortest edge for each pair of boxes as the $\beta$-skeleton edge.
\end{enumerate}

The overall time complexity of this box based $\beta$-skeleton graph construction is $O(n \log n)$, dominated by Delaunay triangulation. There are pathological cases where step 4 will need $O(n^2)$ time, e.g. all the $n$ boxes contain a common overlapping point. But such cases do not happen in OCR results.

\section{Experimental GCN Setup}

The 2-step GCN models are built as in Fig. \ref{fig:5} and Fig. \ref{fig:7}, each carrying 8 steps of graph convolutions with hidden layer size 64 and 4-head self-attention weighted pooling.

At the models' input, each graph node's feature is a vector containing its bounding box information of the word/line. The first five values are width $w$, height $h$, rotation angle $\alpha$, $\cos\alpha$ and $\sin\alpha$. Then for each of its 4 corners $(x_p, y_p)$, we add 6 values $[x_p,$ $x_p\cos\alpha,$ $x_p\sin\alpha,$ $y_p,$ $y_p\cos\alpha,$ $y_p\sin\alpha]$. For line clustering, an additional $w_1$ indicating the first word's width is added to each line for better context of line breaks and list items.

We use cross-entropy loss for both node and edge classification tasks. We train the models from scratch using Momentum optimizer with batch size of 16. The learning rate is set to 0.0002 with a warm-up proportion of 0.01. The training is conducted on 8 Tesla P100 GPUs for approximately 10 hours each model.

\section{Randomized Extension Script for Synthetic Data by Web Scraping}

Almost all web pages use vertical spacing to separate paragraphs, and multi-column text is rare. We use randomly picked and randomly parameterized code pieces from the following table to diversity the web page layout styles.

\begin{table}[htb]
\footnotesize
\caption{Sample web script code for changing paragraph styles.}
\centering
\label{tab:1}       
\begin{tabular}{ll}
\noalign{\smallskip}\noalign{\smallskip}
\hline \hline\noalign{\smallskip}
Style Change & Script Sample \\
\noalign{\smallskip}\hline\noalign{\smallskip}
Single-column to & div.style.columnCount = 2; \\
double-column \\
\noalign{\smallskip}\noalign{\smallskip}
Vertical spacing to & div.style.textIndent = 30px; \\
indentation & div.style.marginTop = 0; \\
& div.style.marginBottom = 0; \\
\noalign{\smallskip}\noalign{\smallskip}
Typography alignment & div.style.textAlign = ``right''; \\
\noalign{\smallskip}\noalign{\smallskip}
Text column width & div.style.width = 50\%; \\
\noalign{\smallskip}\noalign{\smallskip}
Horizontal text block & div.style.marginLeft = 20\%; \\
position \\
\noalign{\smallskip}\noalign{\smallskip}
Line height/spacing & div.style.lineHeight =  150\%; \\
\noalign{\smallskip}\noalign{\smallskip}
Font & div.style.fontFamily = ``times''; \\
\noalign{\smallskip}\hline
\end{tabular}
\end{table}

\end{document}